\documentclass[11pt,a4paper]{article}
\usepackage[usenames,dvipsnames,table]{xcolor}
\usepackage{colortbl}
\usepackage[hyperref]{acl2019}
\usepackage{times}
\usepackage{latexsym}

\usepackage{graphicx}

\usepackage{url}
\usepackage{booktabs}
\usepackage{array}
\usepackage[T1]{fontenc}

\definecolor{orange}{rgb}{1,0.5,0}
\definecolor{lightsalmonpink}{rgb}{1.0, 0.6, 0.6}
\definecolor{verylightsalmonpink}{rgb}{0.966, 0.805, 0.797}
\definecolor{lightblue}{rgb}{0.862, 0.906, 0.984}
\definecolor{lightyellow}{rgb}{1.0, 0.945, 0.797}
\definecolor{lightgreen}{rgb}{0.835, 0.91, 0.828}
\definecolor{lightpurple}{rgb}{0.879, 0.832, 0.902}

\definecolor{lightpink}{rgb}{1.0, 0.71, 0.76}
\aclfinalcopy 
\def\aclpaperid{768} 

\newcommand\customfont[1]{{\usefont{T1}{jd}{m}{n} #1 }}

\newcommand{\namedref}[2]{\hyperref[#2]{#1~\ref*{#2}}}

\newcommand{\sectionref}[1]{\namedref{Section}{sec:#1}}
\newcommand{\tableref}[1]{\namedref{Table}{tab:#1}}
\newcommand{\figureref}[1]{\namedref{Figure}{fig:#1}}
\newcommand{\appendixref}[1]{\namedref{Appendix}{appendix:#1}}

\newcommand{\general}{\textsc{general}}
\newcommand{\specific}{\textsc{specific}}
\newcommand{\yesno}{\textsc{yes-no}}
\newcommand{\squash}{\customfont{SQUASH}}
\newcommand{\question}{\noindent\textbf{Q:}}
\newcommand{\answer}{\noindent\textbf{A:}}
\newcommand{\answernum}[1]{\noindent\textbf{A{#1}:}}

\title{Generating Question-Answer Hierarchies}

\author{Kalpesh Krishna \& Mohit Iyyer \\
  College of Information and Computer Sciences \\
  University of Massachusetts Amherst \\
  {\tt \{kalpesh,miyyer\}@cs.umass.edu} \\\\
  \url{http://squash.cs.umass.edu/} }

\date{}

\begin{document}
\maketitle

\begin{abstract}
The process of knowledge acquisition can be viewed as a question-answer game between a student and a teacher in which the student typically starts by asking broad, open-ended questions before drilling down into  specifics~\cite{hintikka1981logic, hakkarainen2002interrogative}. This pedagogical perspective motivates a new way of representing documents. In this paper, we present \squash~(\textbf{S}pecificity-controlled \textbf{Qu}estion-\textbf{A}n\textbf{s}wer \textbf{H}ierarchies), a novel and challenging text generation task that converts an input document into a hierarchy of question-answer pairs. Users can click on high-level questions (e.g., ``Why did Frodo leave the Fellowship?'') to reveal related but more specific questions (e.g., ``Who did Frodo leave with?''). Using a question taxonomy loosely based on~\newcite{lehnert1978process}, we classify questions in existing reading comprehension datasets as either \general~or \specific. We then use these labels as input to a pipelined system centered around a conditional neural language model. We extensively evaluate the quality of the generated QA hierarchies through crowdsourced experiments and report strong empirical results.
\end{abstract}
\section{Introduction}
\label{sec:introduction}

\begin{figure}[t!]
\centering
\includegraphics[scale=0.97]{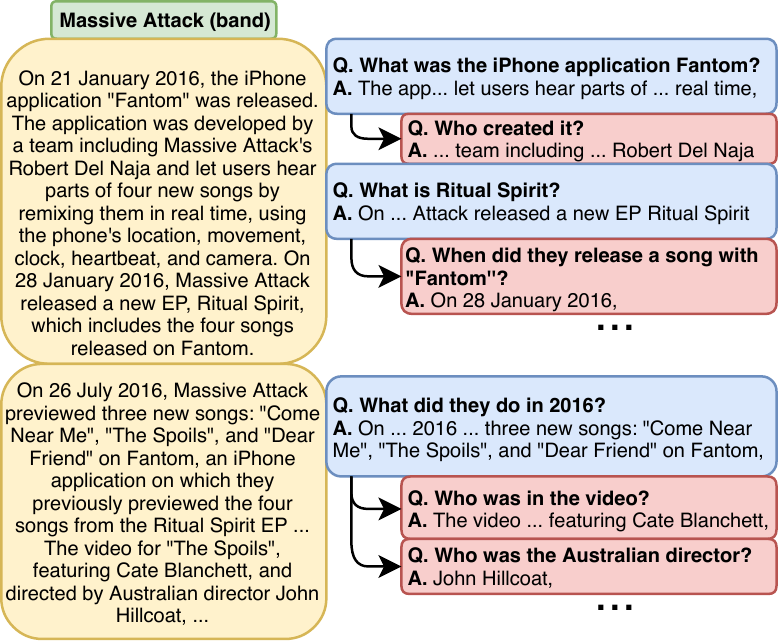}
\caption{A subset of the QA hierarchy generated by our \squash\ system that consists of \colorbox{lightblue}{\general}~and \colorbox{verylightsalmonpink}{\specific}~questions with extractive answers.}
\label{fig:sample}
\vspace{-0.1in}
\end{figure}

\question~\textit{What is this paper about?}\\
\answer~We present a novel text generation task which converts an input document into a model-generated hierarchy of question-answer (QA) pairs arranged in a top-down tree structure (\figureref{sample}). Questions at higher levels of the tree are broad and open-ended while questions at lower levels ask about more specific factoids. 
An entire document has multiple root nodes (``key ideas'') that unfold into a forest of question trees. While readers are initially shown only the root nodes of the question trees, they can ``browse'' the document by clicking on root nodes of interest to reveal more fine-grained related information. We call our task \squash~(\textbf{S}pecificity-controlled \textbf{Qu}estion \textbf{A}n\textbf{s}wer \textbf{H}ierarchies).\\

\vspace{-0.1in}

\question~\textit{Why represent a document with QA pairs?}\footnote{Our introduction is itself an example of the QA format. Other academic papers such as ~\newcite{Henderson2018DeepRL} have also used this format to effectively present information.}\\
\answer~ Questions and answers (QA) play a critical role in scientific inquiry, information-seeking dialogue and knowledge acquisition~\cite{hintikka1981logic, hintikka1988logic, stede2004information}. For example, web users often use QA pairs to manage and share knowledge~\citep{wagner2004wiki, wagner2005supporting, gruber2008collective}. Additionally, unstructured lists of ``frequently asked questions'' (FAQs) are regularly deployed at scale to present information. Industry studies have demonstrated their effectiveness at cutting costs associated with answering customer calls or hiring technical experts~\cite{davenport1998successful}. Automating the generation of QA pairs can thus be of immense value to companies and web communities.

\vspace{-0.1in}

\question~\textit{Why add hierarchical structure to QA pairs?}\\
\answer~
While unstructured FAQs are useful, pedagogical applications benefit from additional hierarchical organization. \newcite{hakkarainen2002interrogative} show that students learn concepts effectively by first asking general, explanation-seeking questions before drilling down into more specific questions. More generally, hierarchies break up content into smaller, more digestable chunks. User studies demonstrate a strong preference for hierarchies in document summarization~\cite{buyukkokten2001seeing, christensen2014hierarchical} since they help readers easily identify and explore key topics~\cite{zhang2017}.\\

\vspace{-0.1in}

\question~\textit{How do we build systems for \squash?}\\
\answer~We leverage the abundance of reading comprehension QA datasets to train a pipelined system for \squash. One major challenge is the lack of labeled hierarchical structure within existing QA datasets; we tackle this issue in~\sectionref{classification} by using the question taxonomy of~\newcite{lehnert1978process} to classify questions in these datasets as either \general\ or \specific. We then condition a neural question generation system on these two classes, which enables us to generate both types of questions from a paragraph. We filter and structure these outputs using the techniques described in \sectionref{model}.\\

\vspace{-0.1in}

\question~\textit{How do we evaluate our \squash\ pipeline?}\\
\answer~
Our crowdsourced evaluation (\sectionref{evaluation}) focuses on fundamental properties of the generated output such as QA quality, relevance, and hierarchical correctness. Our work is a first step towards integrating QA generation into document understanding; as such, we do not directly evaluate how \emph{useful} \squash\ output is for downstream pedagogical applications. Instead, a detailed qualitative analysis (\sectionref{qual}) identifies challenges that need to be addressed before \squash\ can be deployed to real users.\\

\vspace{-0.1in}

\question~\textit{What are our main contributions?}\\
\answernum{1} A method to classify questions according to their specificity based on ~\newcite{lehnert1978process}.\\
\answernum{2} A model controlling specificity of generated questions, unlike prior work on QA generation.\\
\answernum{3} A novel text generation task (\squash), which converts documents into specificity-based hierarchies of QA pairs.\\
\answernum{4} A pipelined system to tackle \squash~along with crowdsourced methods to evaluate it.


\question~\textit{How can the community build on this work?}\\
\answer~
We have released our codebase, dataset and a live demonstration of our system at \url{http://squash.cs.umass.edu/}. Additionally, we outline guidelines for future work in \sectionref{future}.

\section{Obtaining training data for \squash}
\label{sec:classification}

The proliferation of reading comprehension datasets like SQuAD~\cite{rajpurkar2016squad, Rajpurkar2018KnowWY} has enabled state-of-the-art neural question generation systems~\cite{du2017learning, kim2018improving}. However, these systems are trained for \emph{individual} question generation, while the goal of \squash\ is to produce a general-to-specific hierarchy of QA pairs. Recently-released conversational QA datasets like QuAC~\cite{ChoiQuAC2018} and CoQA~\cite{reddy2018coqa} contain a \emph{sequential} arrangement of QA pairs, but question specificity is not explicitly marked.\footnote{``Teachers'' in the QuAC set-up can encourage ``students'' to ask a follow-up question, but we cannot use these annotations to infer a hierarchy because students are not required to actually follow their teachers' directions.}
Motivated by the lack of hierarchical QA datasets, we automatically classify questions in SQuAD, QuAC and CoQA according to their specificity using a combination of rule-based and automatic approaches.

\begin{table*}
\small
\begin{center}
\begin{tabular}{ p{4.6cm}p{1.5cm}p{3.5cm}p{4.3cm} } 
 \toprule
 \bf Conceptual class & \bf Specificity & \bf Question asks for... & \bf Sample templates \\
 \midrule
 Causal Antecedent, Goal Oriented, Enablement, Causal Consequent, Expectational & \general & the reason for occurrence of an event and the consequences of it & \emph{Why ...}, \emph{What happened after / before ...}, \emph{What was the cause / reason / purpose ...},  \emph{What led to ...}\\
 \midrule
 Instrumental \newline & \general & a procedure / mechanism & \emph{How} question with VERB parent for \emph{How} in dependency tree\\
 \midrule
 Judgemental & \general & a listener's opinion &  Words like \emph{you}, \emph{your} present \\
 \midrule
 Concept Completion, Feature Specification & \general~~\emph{or} \newline \specific & fill-in-the-blank information & \emph{Where / When / Who ...} \newline (``\specific'' templates) \\
 \midrule
 Quantification & \specific & an amount & \emph{How many / long ...} \\
  \midrule
   Verification, Disjunctive & \yesno & Yes-No answers & first word is VERB \\
 \midrule
 Request & N/A & an act to be performed & (absent in datasets) \\
 \bottomrule
\end{tabular}
\end{center}
\caption{The 13 conceptual categories of~\newcite{lehnert1978process} and some templates to identify them and their specificity.}
\vspace{-0.2in}
\label{tab:templates}
\end{table*}

\subsection{Rules for specificity classification}
What makes one question more specific than another? Our scheme for classifying question specificity maps each of the 13 conceptual question categories defined by~\newcite{lehnert1978process} to three coarser labels: \general, \specific, or \yesno.\footnote{We add a third category for \yesno~questions as they are difficult to classify as either \general~or \specific.} As a result of this mapping, \specific~questions usually ask for low-level information (e.g., entities or numerics), while \general~questions ask for broader overviews (e.g., ``what happened in 1999?'') or causal information (e.g, ``why did...''). Many question categories can be reliably identified using simple templates and rules; A complete list is provided in \tableref{templates}.\footnote{Questions in~\newcite{lehnert1978process} were classified using a conceptual dependency parser~\cite{schank1972conceptual}. We could not find a modern implementation of this parser and thus decided to use a rule-based approach that relies on spaCy 2.0~\citep{spacy2} for all preprocessing.} \\

\vspace{-0.1in}

\noindent\textbf{Classifying questions not covered by templates: }
If a question does not satisfy any template or rule, how do we assign it a label? 
We manage to classify roughly half of all questions with our templates and rules (\tableref{scheme_distro}); for the remaining half, we resort to a data-driven approach. First, we manually label 1000 questions in QuAC\footnote{We use QuAC because its design encourages a higher percentage of \general~questions than other datasets, as the question-asker was unable to read the document to formulate more specific questions.} using our specificity labels. This annotated data is then fed to a single-layer CNN binary classifier~\cite{kim2014convolutional} using ELMo contextualized embeddings~\cite{peters2018deep}.\footnote{Implemented in AllenNLP~\cite{gardner2018allennlp}.} On a 85\%-15\% train-validation split, we achieve a high classification accuracy of 91\%.  The classifier also transfers to other datasets: on 100 manually labeled CoQA questions, we achieve a classification accuracy of 80\%. To obtain our final dataset (\tableref{class_distro}), we run our rule-based approach on all questions in SQuAD 2.0, QuAC, and CoQA and apply our classifier to label questions that were not covered by the rules.
We further evaluate the specificity of the questions generated by our final system using a crowdsourced study in \sectionref{structural}.
\begin{table}[h!]
\small
\begin{center}
\begin{tabular}{ lrrrr } 
 \toprule
Dataset & Size & \general & \specific & \yesno \\ 
\midrule
SQuAD & 86.8k & 28.2\% & 69.7\% & 2.1\% \\
QuAC & 65.2k & 34.9\% & 33.5\% & 31.6\% \\
CoQA & 105.6k & 23.6\% & 54.9\% & 21.5\% \\
All & 257.6k & 28.0\% & 54.5\% & 17.5\% \\
\bottomrule
\end{tabular}
\end{center}
\caption{Distribution of classes in the final datasets. We add some analysis on this distribution in \appendixref{classification}.}
\label{tab:class_distro}
\vspace{-0.2in}
\end{table}
 \begin{figure*}[t!]
\centering
\includegraphics[scale=0.8]{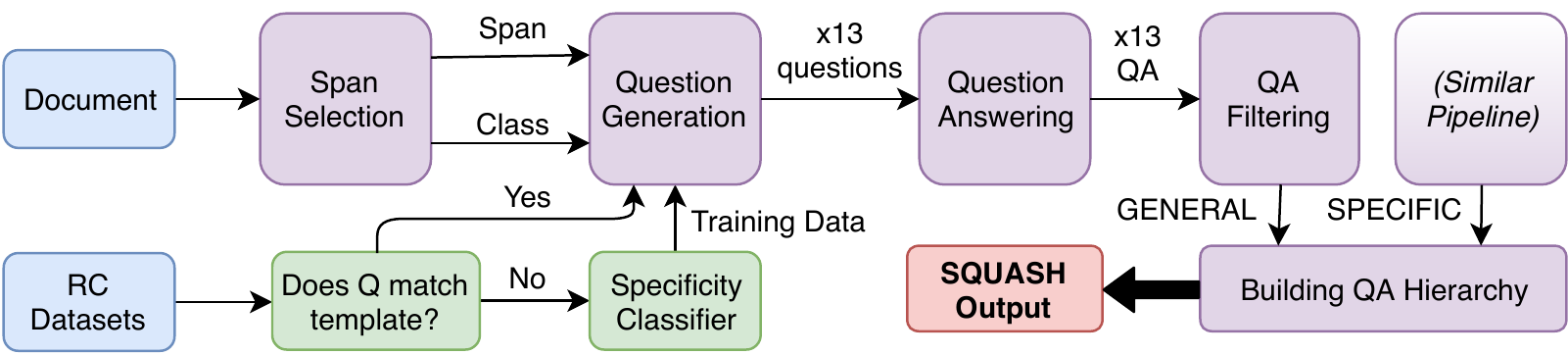}
\caption{An overview of the process by which we generate \colorbox{verylightsalmonpink}{a pair of \general-\specific~questions}, which consists of feeding \colorbox{lightblue}{input data} (\emph{``RC''} is Reading Comprehension) through various modules, including a \colorbox{lightgreen}{question classifier} and a multi-stage \colorbox{lightpurple}{pipeline} for question generation, answering, and filtering.}
\label{fig:pipeline}
\vspace{-0.15in}
\end{figure*}

\section{A pipeline for \squash ing documents}
\label{sec:model}
To \squash\ documents, we build a pipelined system (\figureref{pipeline})  that takes a single paragraph as input and produces a hierarchy of QA pairs as output; for multi-paragraph documents, we \squash\ each paragraph independently of the rest. At a high level, the pipeline consists of five steps: (1) answer span selection, (2) question generation conditioned on answer spans and specificity labels, (3) \emph{extractively} answering generated questions, (4) filtering out bad QA pairs, and (5) structuring the remaining pairs into a \general-to-\specific\ hierarchy. The remainder of this section describes each step in more detail and afterwards explains how we leverage pretrained language models  to improve individual components of the pipeline.

\subsection{Answer span selection}
\label{sec:model_span}
Our pipeline begins by selecting an answer span from which to generate a question. To train the system, we can use ground-truth answer spans from our labeled datasets, but at test time how do we select answer spans? Our solution is to consider all individual sentences in the input paragraph as potential answer spans (to generate \general~and \specific~questions), along with all entities and numerics (for just \specific~ questions). We did not use data-driven sequence tagging approaches like previous work~\citep{du2017identifying,Du2018HarvestingPQ}, since our preliminary experiments with such approaches yielded poor results on QuAC.\footnote{We hypothesize that answer span identification on QuAC is difficult because the task design encouraged ``teachers'' to provide more information than just the minimal answer span.} More details are provided in \appendixref{notwork}.


\subsection{Conditional question generation}
Given a paragraph, answer span, and desired specificity label, we train a neural encoder-decoder model on all three reading comprehension datasets (SQuAD, QuAC and CoQA) to generate an appropriate question.\\

\vspace{-0.1in}

\noindent\textbf{Data preprocessing: }
At training time, we use the ground-truth answer spans from these datasets as input to the question generator. To improve the quality of \specific~questions generated from sentence spans, we use the extractive \emph{evidence} spans for CoQA instances~\cite{reddy2018coqa} instead of the shorter, partially abstractive answer spans~\cite{yatskar2018qualitative}.  In all datasets, we remove unanswerable questions and questions whose answers span multiple paragraphs. A few very generic questions (e.g. ``what happened in this article?'') were manually identified removed from the training dataset. Some other questions (e.g., ``where was he born?'') are duplicated many times in the dataset; we downsample such questions to a maximum limit of 10. Finally, we preprocess both paragraphs and questions using byte-pair encoding~\cite{sennrich2016neural}.\\
 
 \vspace{-0.1in}
 
\noindent\textbf{Architecture details: }
We use a two-layer biLSTM encoder and a single-layer LSTM~\citep{hochreiter1997long} decoder with soft attention~\citep{Bahdanau2014NeuralMT} to generate questions, similar to~\citet{du2017learning}. Our architecture is augmented with a copy mechanism~\cite{see2017get} over the encoded paragraph representations. Answer spans are marked with \texttt{<SOA>} and \texttt{<EOA>} tokens in the paragraph, and representations for tokens within the answer span are attended to by a separate attention head. We condition the decoder on the specificity class (\general, \specific~and \yesno)\footnote{While we do not use \yesno~questions at test time, we keep this class to avoid losing a significant proportion of training data.} by concatenating an embedding for the ground-truth class to the input of each time step. We implement models in PyTorch \texttt{v0.4}~\cite{paszke2017automatic}, and the best-performing model achieves a perplexity of 11.1 on the validation set. Other hyperparameters details are provided in \appendixref{hyperparameters}. \\

\vspace{-0.1in}

\noindent\textbf{Test time usage: }
At test time, the question generation module is supplied with answer spans and class labels as described in \sectionref{model_span}. To promote diversity, we over-generate prospective candidates~\cite{heilman2010good} for every answer span and later prune them. Specifically, we use beam search with a beam size of 3 to generate three highly-probable question candidates. As these candidates are often generic, we additionally use \emph{top-k random sampling}~\cite{Fan2018HierarchicalNS} with $k=10$, a recently-proposed diversity-promoting decoding algorithm, to generate ten more question candidates per answer span. Hence, for every answer span we generate \textbf{13} question candidates. We discuss issues with using just standard beam search for question generation in \sectionref{workswell}.

\subsection{Answering generated questions}
While we condition our question generation model on pre-selected answer spans, the generated questions may not always correspond to these input spans. Sometimes, the generated questions are either unanswerable or answered by a different span in the paragraph. By running a pretrained QA model over the generated questions, we can detect questions whose answers do not match their original input spans and filter them out. The predicted answer for many questions has partial overlap with the original answer span; in these cases, we display the predicted answer span during evaluation, as a qualitative inspection shows that the predicted answer is more often closer to the correct answer. For all of our experiments, we use the AllenNLP implementation of the BiDAF++ question answering model of~\newcite{ChoiQuAC2018} trained on QuAC with no dialog context.

\subsection{Question filtering}
\label{sec:model_filter}
After over-generating candidate questions from a single answer span, we use simple heuristics to filter out low-quality QA pairs. We remove generic and duplicate question candidates\footnote{Running \emph{Top-k random sampling} multiple times can produce duplicate candidates, including those already in the top beams.} and pass the remaining QA pairs through the multi-stage question filtering process described below.\\

\vspace{-0.1in}

\noindent\textbf{Irrelevant or repeated entities:} \emph{Top-k random sampling} often generates irrelevant questions; we reduce their incidence by removing any candidates that contain nouns or entities unspecified in the passage. As with other neural text generation systems~\cite{Holtzman2018LearningTW}, we commonly observe repetition in the generated questions and deal with this phenomenon by  removing candidates with repeated nouns or entities.\\

\vspace{-0.1in}

\noindent\textbf{Unanswerable or low answer overlap:} We remove all candidates marked as ``unanswerable'' by the question answering model, which prunes 39.3\% of non-duplicate question candidates. These candidates are generally grammatically correct but considered irrelevant to the original paragraph by the question answering model. Next, we compute the overlap between original and predicted answer span by computing word-level precision and recall~\cite{rajpurkar2016squad}. For \general~questions generated from sentence spans, we attempt to maximize recall by setting a minimum recall threshold of 0.3.\footnote{Minimum thresholds were qualitatively chosen based on the specificity type. } Similarly, we maximize recall for \specific~questions generated from named entities with a minimum recall constraint of 0.8. Finally, for \specific~questions generated from sentence spans, we set a minimum \emph{precision} threshold of 1.0, which filters out questions whose answers are not completely present in the ground-truth sentence.\\

\vspace{-0.1in}

\noindent\textbf{Low generation probability:} If multiple candidates remain after applying the above filtering criteria, we select the most probable candidate for each answer span. \specific~questions generated from sentences are an exception to this rule: for these questions, we select the ten most probable candidates, as there might be multiple question-worthy bits of information in a single sentence.
If no candidates remain, in some cases\footnote{For example, if no valid \general~questions for the entire paragraph are generated.} we use a fallback mechanism that sequentially ignores filters to retain more candidates. 

\vspace{-0.05in}

\subsection{Forming a QA hierarchy}
\label{sec:grouping}
\begin{figure}[t!]
\centering
\includegraphics[scale=0.9]{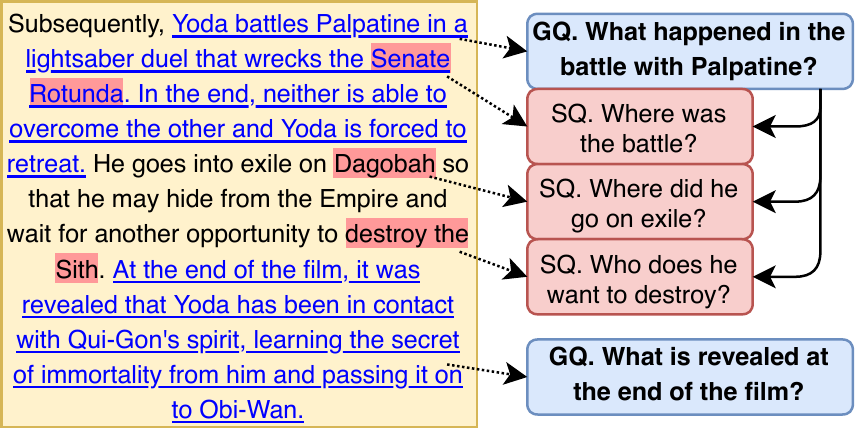}
\caption{Procedure used to form a QA hierarchy. The predicted answers for GQs (\colorbox{lightblue}{\general}~questions), are underlined in \textcolor{blue}{\underline{blue}}. The predicted answers for SQs (\colorbox{verylightsalmonpink}{\specific}~questions) are highlighted in \colorbox{lightsalmonpink}{red}.}
\label{fig:overlap}
\vspace{-0.15in}
\end{figure}

The output of the filtering module is an unstructured list of \general~and \specific~QA pairs generated from a single paragraph. \figureref{overlap} shows how we group these questions into a meaningful hierarchy. First, we choose a parent for each \specific~question by maximizing the overlap (word-level precision) of its predicted answer with the predicted answer for every \general~question. If a \specific~question's answer does not overlap with any \general~question's answer (e.g., \emph{``Dagobah''} and \emph{``destroy the Sith''}) we map it to the closest \general~question whose answer occurs \emph{before} the \specific~question's answer (\emph{``What happened in the battle ...?''}).\footnote{This heuristic is justified because users read \general~questions \emph{before} \specific~ones in our interface.}

\subsection{Leveraging pretrained language models}
Recently, pretrained language models based on the Transformer architecture~\cite{Vaswani2017AttentionIA} have significantly boosted question answering performance~\cite{devlin2019bert} as well as the quality of conditional text generation~\cite{wolf2019}.
Motivated by these results, we modify components of the pipeline to incorporate language model pretraining for our demo. Specifically, our demo's question answering module is the BERT-based model in ~\newcite{devlin2019bert}, and the question generation module is trained by fine-tuning the publicly-available GPT2-small model~\cite{radford2019language}. Please refer to~\appendixref{modifications} for more details. These modifications produce better results qualitatively and speed up the \squash~pipeline since question overgeneration is no longer needed.

Note that the figures and results in \sectionref{evaluation} are using the original components described above.

\begin{table*}[ht!]
\begin{center}
\begin{tabular}{ lcccc } 
 \toprule
 \textbf{Experiment} & \multicolumn{2}{c}{\underline{\textbf{Generated}}} & \multicolumn{2}{c}{\underline{\textbf{Gold}}} \\
 & Score & Fleiss~$\kappa$ & Score & Fleiss~$\kappa$ \\
 \midrule
 Is this question well-formed? & 85.8\% & 0.65 & 93.3\% & 0.54 \\
 \midrule
 Is this question relevant? & 78.7\% & 0.36 & 83.3\% & 0.41 \\
 ~~~~(among \emph{well-formed}) & 81.1\% & 0.39 & 83.3\% & 0.40 \\
 \midrule
  Does the span \emph{partially} contain the answer? & 85.3\% & 0.45 & 81.1\% & 0.43 \\
  ~~~~(among \emph{well-formed}) & 87.6\% & 0.48 & 82.1\% & 0.42 \\
  ~~~~(among \emph{well-formed and relevant}) & 94.9\% & 0.41 & 92.9\% & 0.44 \\
 \midrule
 Does the span \emph{completely} contain the answer? & 74.1\% & 0.36 & 70.0\% & 0.37 \\
  ~~~~(among \emph{well-formed}) & 76.9\% & 0.36 & 70.2\% & 0.39\\
    ~~~~(among \emph{well-formed and relevant}) & 85.4\% & 0.30 & 80.0\% & 0.42 \\
 \bottomrule
\end{tabular}
\end{center}
\caption{Human evaluations demonstrate the high individual QA quality of our pipeline's outputs. All interannotator agreement scores (Fleiss $\kappa$) show ``fair'' to ``substantial'' agreement~\cite{landis1977measurement}.}
\label{tab:evaluation}
\end{table*}
\section{Evaluation}
\label{sec:evaluation}

We evaluate our \squash\ pipeline on documents from the QuAC development set using a variety of crowdsourced\footnote{All our crowdsourced experiments were conducted on the Figure Eight platform with three annotators per example (scores calculated by counting examples with two or more correct judgments). We hired annotators from predominantly English-speaking countries with a rating of at least Level 2, and we paid them between 3 and 4 cents per judgment.} experiments. Concretely, we evaluate the quality and relevance of individual questions, the relationship between generated questions and predicted answers, and the structural properties of the QA hierarchy.
We emphasize that our experiments examine only the \emph{quality} of a \squash ed document, not its actual \emph{usefulness} to downstream users. Evaluating usefulness (e.g., measuring if \squash\ is more helpful than the input document) requires systematic and targeted human studies~\citep{buyukkokten2001seeing} that are beyond the scope of this work. 

\subsection{Individual question quality and relevance}
\label{sec:eval_question}
Our first evaluation measures whether questions generated by our system are \emph{well-formed} (i.e., grammatical and pragmatic). We ask crowd workers whether or not a given question is both grammatical and meaningful.\footnote{As ``meaningful'' is potentially a confusing term for crowd workers, we ran another experiment asking only for grammatical correctness and achieved very similar results.} For this evaluation, we acquire judgments for 200 generated QA pairs and 100 gold QA pairs\footnote{Results on this experiment were computed after removing 3 duplicate generated questions and 10 duplicate gold questions.} from the QuAC validation set (with an equal split between \general~and \specific~questions). The first row of \tableref{evaluation} shows that 85.8\% of generated questions satisfy this criterion with a high agreement across workers. 

\noindent\textbf{Question relevance: }
How many generated questions are actually relevant to the input paragraph? While the percentage of \emph{unanswerable} questions that were generated offers some insight into this question, we removed all of them during the filtering pipeline (\sectionref{model_filter}). Hence, we display an input paragraph and generated question to crowd workers (using the same data as the previous well-formedness evaluation) and ask whether or not the paragraph contains the answer to the question. The second row of \tableref{evaluation} shows that 78.7\% of our questions are relevant to the paragraph, compared to 83.3\% of gold questions. 

\begin{figure*}[t]
\centering
\includegraphics[scale=1]{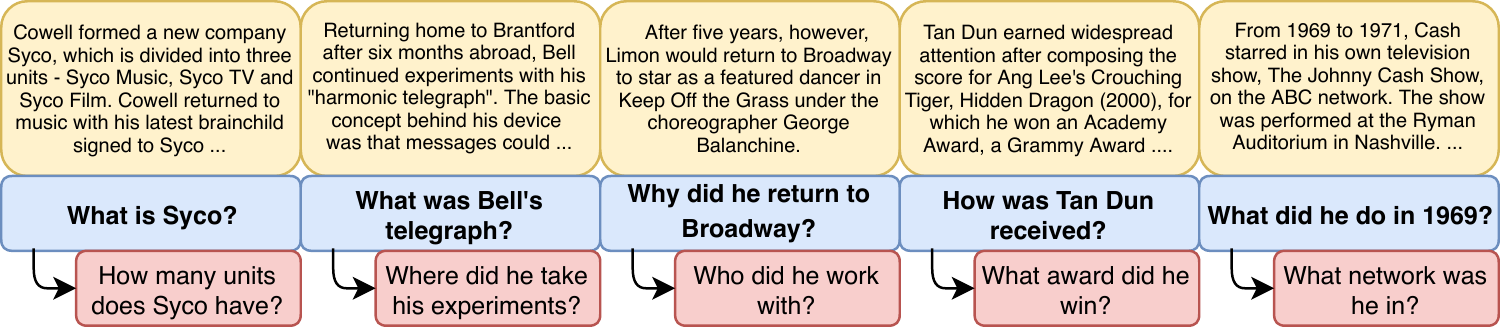}
\caption{\squash~question hierarchies generated by our system with \colorbox{lightyellow}{reference snippets}. Questions in the hierarchy are of the correct specificity class (i.e., \colorbox{lightblue}{\general}, \colorbox{verylightsalmonpink}{\specific}).}
\label{fig:example}
\vspace{-0.2in}
\end{figure*}

\subsection{Individual answer validity}
\label{sec:eval_answer}
Is the predicted answer actually a valid answer to the generated question? 
In our filtering process, we automatically measured answer overlap between the input answer span and the predicted answer span and used the results to remove low-overlap QA pairs. To evaluate answer recall after filtering, we perform a crowdsourced evaluation on the same 300 QA pairs as above by asking crowdworkers whether or not a predicted answer span \emph{contains} the answer to the question. We also experiment with a more relaxed variant (\emph{partially contains} instead of \emph{completely contains}) and report results for both task designs in the third and fourth rows of \tableref{evaluation}. Over 85\% of predicted spans partially contain the answer to the generated question, and this number increases if we consider only questions that were previously labeled as \emph{well-formed} and \emph{relevant}. The lower gold performance is due to the contextual nature of the gold QA pairs in QuAC, which causes some questions to be meaningless in isolation (e.g.\emph{``What did she do next?''} has unresolvable coreferences).

\begin{table}[h!]
\small
\begin{center}
\begin{tabular}{ p{4cm}cc } 
 \toprule
 \textbf{Experiment} & \textbf{Score} & \textbf{Fleiss~$\kappa$} \\
 \midrule
Which question type asks for more information? & 89.5\% & 0.57 \\
 \midrule
 Which \specific~question is closer to \general~QA? & & \\ ~~~~~~~~\emph{different paragraph} & 77.0\% & 0.47 \\
 ~~~~~~~~\emph{same paragraph} & 64.0\% & 0.30 \\
 \bottomrule
\end{tabular}
\end{center}
\vspace{-0.09in}
\caption{Human evaluation of the structural correctness of our system. The labels \emph{``different / same paragraph''} refer to the location of the \emph{intruder} question. The results show the accuracy of specificity and hierarchies.}
\label{tab:evaluation_structure}
\vspace{-0.1in}
\end{table}

\vspace{-0.1in}

\subsection{Structural correctness}
\label{sec:structural}
To examine the hierachical structure of \squash ed documents, we conduct three experiments.\\

\vspace{-0.1in}

\noindent\textbf{How faithful are output questions to input specificity?}
First, we investigate whether our model is actually generating questions with the correct specificity label. We run our specificity classifier (\sectionref{classification}) over 400 randomly sampled questions (50\% \general, 50\% \specific) and obtain a high classification accuracy of 91\%.\footnote{Accuracy computed after removing 19 duplicates.} This automatic evaluation suggests the model is capable of generating different types of questions.\\

\vspace{-0.1in}

\noindent\textbf{Are \general~questions more representative of a paragraph than \specific~questions?}
To see if \general~questions really do provide more high-level information, we sample 200 \general-\specific~question pairs\footnote{We avoid gold-standard control experiments for structural correctness tests since questions in the QuAC dataset were not generated with a hierarchical structure in mind. Pilot studies using our question grouping module on gold data led to sparse hierarchical structures which were not favored by our crowd workers.} grouped together as described in~\sectionref{grouping}. For each pair of questions (without showing answers), we ask crowd workers to choose the question which, if answered, would give them more information about the paragraph. As shown in \tableref{evaluation_structure}, in 89.5\% instances the \general~question is preferred over the \specific~one, which confirms the strength of our specificity-controlled question generation system.\footnote{We also ran a pilot study asking workers \emph{``Which question has a longer answer?''} and observed a higher preference of 98.6\% for \general~questions.}\\

\vspace{-0.1in}

\noindent\textbf{How related are \specific~questions to their parent \general~question?}
Finally, we investigate the effectiveness of our question grouping strategy, which bins multiple \specific~QA pairs under a single \general~QA pair. We show crowd workers a reference \general~QA pair and ask them to choose the most related \specific~ question given two choices, one of which is the system's output and the other an \emph{intruder} question. We randomly select intruder \specific~questions from either a different paragraph within the same document or a different group within the same paragraph. As shown in \tableref{evaluation_structure}, crowd workers prefer the system's generated \specific~question with higher than random chance (50\%) regardless of where the intruder comes from. As expected, the preference and agreement is higher when intruder questions come from different paragraphs, since groups within the same paragraph often contain related information (\sectionref{shortcomings}).
\begin{figure}[t!]
\centering
\includegraphics[scale=0.7]{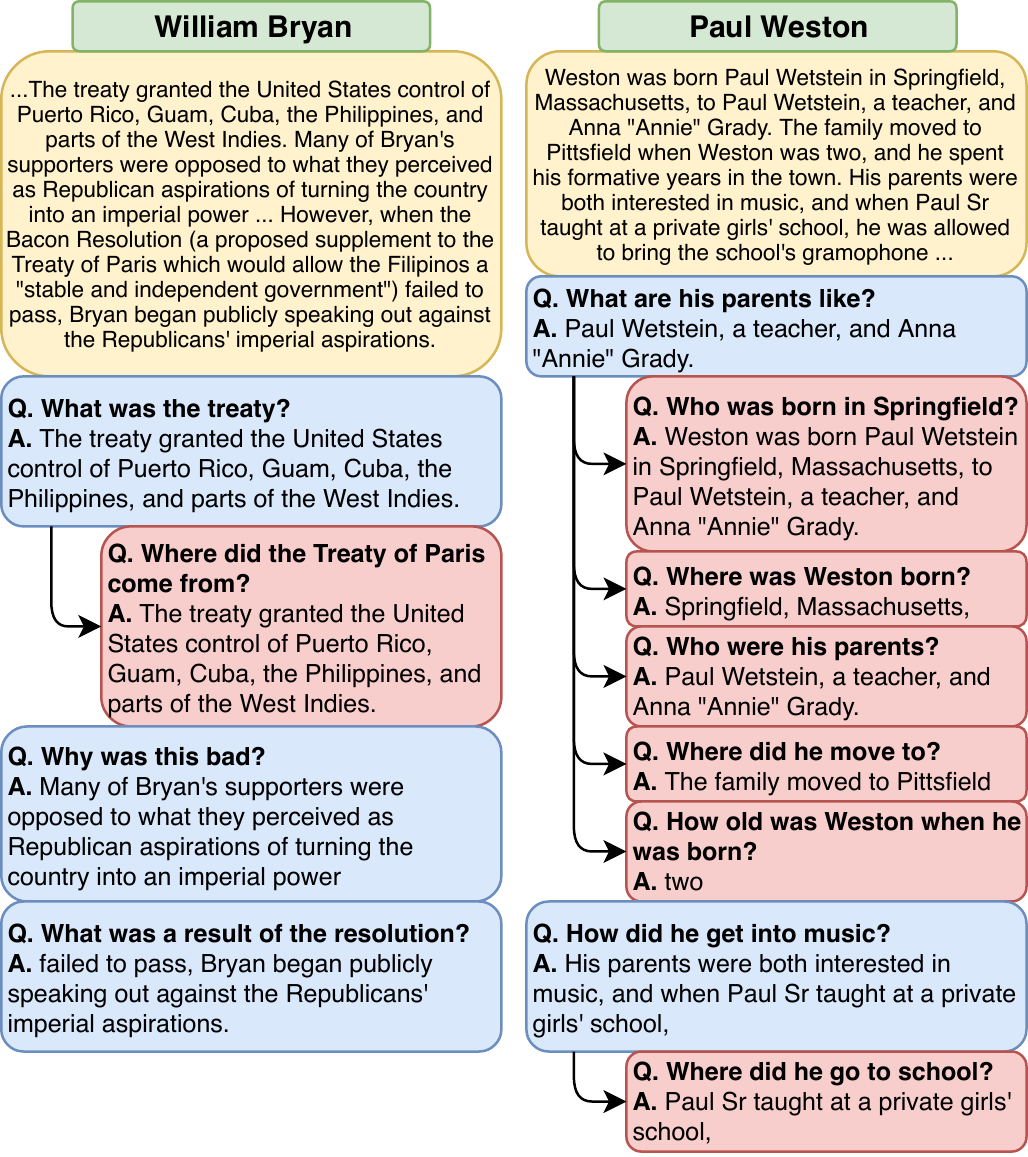}
\caption{Two \squash~outputs generated by our system. The \colorbox{lightgreen}{William Bryan} example has interesting \colorbox{lightblue}{\general}~questions. The \colorbox{lightgreen}{Paul Weston} example showcases several mistakes our model makes.}
\label{fig:massexamples}
\end{figure}

\section{Qualitative Analysis}
\label{sec:qual}
In this section we analyze outputs (\figureref{example}, \figureref{massexamples}) of our pipeline and identify its strengths and weaknesses. We additionally provide more examples in the appendix (\figureref{mass_examples}).

\vspace{-0.05in}

\subsection{What is our pipeline good at?}
\label{sec:workswell}
\paragraph{Meaningful hierarchies:} Our method of grouping the generated questions (\sectionref{grouping}) produces hierarchies that clearly distinguish between \general\ and \specific\ questions;~\figureref{example} contains some hierarchies that support the positive results of our crowdsourced evaluation.

\vspace{-0.05in}

\paragraph{\emph{Top-k} sampling:}
Similar to prior work~\cite{Fan2018HierarchicalNS, holtzman2019curious}, we notice that beam search often produces generic or repetitive beams (\tableref{beam}). Even though the \emph{top-k} scheme always produces lower-probable questions than beam search, our filtering system prefers a \emph{top-k} question \textbf{49.5\%}  of the time.

\begin{table}[t]
\begin{center}
\footnotesize
\begin{tabular}{ p{0.3cm}p{6.5cm} } 
 \toprule
 \multicolumn{2}{p{7cm}}{\emph{``In \textbf{1942}, Dodds enlisted in the US army and served as an anti aircraft gunner during World War II.''}}  \\
  \hline
 \rowcolor{verylightsalmonpink}
 ~\newline B \newline & In what year did the US army take place? \newline In what year did the US army take over? \newline  In what year did the US army take place in the US?\\
 \hline
 \rowcolor{lightblue}
~\newline T \newline & What year was he enlisted? \newline When did he go to war? \newline  When did he play as anti aircraft?\\
  \hline
\end{tabular}
\end{center}
\caption{\colorbox{verylightsalmonpink}{Beam Search (B)} vs \colorbox{lightblue}{Top\emph{-k} sampling (T)} for \specific~ question generation. Top-\emph{k} candidates tend to be more diverse.}
\label{tab:beam}
\end{table}

\subsection{What kind of mistakes does it make?}
\label{sec:shortcomings}
We describe the various types of errors our model makes in this section, using the \colorbox{lightgreen}{Paul Weston} \squash~output in \figureref{massexamples} as a running example. Additionally, we list some modeling approaches we tried that did \emph{not} work in \appendixref{notwork}.

\vspace{-0.05in}

\paragraph{Reliance on a flawed answering system:}
Our pipeline's output is tied to the quality of the pretrained answering module, which both filters out questions and produces final answers. QuAC has long answer spans~\cite{ChoiQuAC2018} that cause low-precision predictions with extra information (e.g., \emph{``Who was born in Springfield?''}). Additionally, the answering module occasionally swaps two named entities present in the paragraph.\footnote{For instance in the sentence \emph{``The Carpenter siblings were born in New Haven, to Harold B. and Agnes R.''} the model \emph{incorrectly} answers the question \emph{``Who was born in New Haven?''} as \emph{``Harold B. and Agnes R.''}}

\vspace{-0.05in}

\paragraph{Redundant information and lack of discourse:} In our system, each QA pair is generated independently of all the others. Hence, our outputs lack an inter-question discourse structure. Our system often produces a pair of redundant \specific~questions where the text of one question answers the other (e.g., \emph{``Who was born in Springfield?''} vs. \emph{``Where was Weston born?''}). These errors can likely be corrected by conditioning the generation module on previously-produced questions (or additional filtering); we leave this to future work. 

\vspace{-0.05in}

\paragraph{Lack of world knowledge:} Our models lack commonsense knowledge (\emph{``How old was Weston when he was born?''}) and can misinterpret polysemous words. Integrating pretrained contextualized embeddings~\cite{peters2018deep} into our pipeline is one potential solution.

\vspace{-0.05in}

\paragraph{Multiple \general~QA per paragraph:} Our system often produces more than one tree per paragraph, which is undesirable for short, focused paragraphs with a single topic sentence. To improve the user experience, it might be ideal to restrict the number of \general\ questions we show per paragraph. While we found it difficult to generate \general~questions representative of entire paragraphs (\appendixref{notwork}), a potential solution could involve identifying and generating questions from topic sentences.

\vspace{-0.05in}

\paragraph{Coreferences in \general~questions:} Many generated \general~questions contain coreferences due to contextual nature of the QuAC and CoQA training data (\emph{``How did \underline{\textbf{he}} get into music?''}). Potential solutions could involve either constrained decoding to avoid beams with anaphoric expressions or using the CorefNQG model of~\newcite{Du2018HarvestingPQ}.

\subsection{Which models did not work?}
\label{sec:notwork}
We present modelling approaches which did not work in \appendixref{notwork}. This includes, i) end-to-end modelling to generate sequences of questions using QuAC, ii) span selection NER system, iii) generation of \general~questions representative of entire paragraphs, iv) answering system trained on the combination of QuAC, CoQA and SQuAD.

\section{Related Work}
Our work on \squash~is related to research in
three broad areas: question generation, information
retrieval and summarization.

\vspace{-0.05in}

\paragraph{Question Generation:} Our work builds upon neural question generation systems~\cite{du2017learning,Du2018HarvestingPQ}. Our work conditions generation on specificity, similar to difficulty-conditioned question generation~\cite{gao2018difficulty}. QA pair generation has previously been used for dataset creation~\cite{serban2016generating,Du2018HarvestingPQ}. Joint modeling of question generation and answering has improved the performance of individual components~\cite{tang2017question, wang2017joint, sachan2018self} and enabled visual dialog generation~\cite{jain2018two}.

\vspace{-0.05in}

\paragraph{Information Retrieval:} Our hierarchies are related to interactive retrieval setting~\cite{hardtke2009demonstration, brandt2011dynamic} where similar webpages are grouped together. \squash~is also related to exploratory~\cite{marchionini2006exploratory} and faceted search~\cite{yee2003faceted}.

\vspace{-0.05in}

\paragraph{Summarization:} Our work is related to \textit{query-focused} summarization~\cite{dang2005overview,baumel2018query} which conditions an output summary on an input query. Hierarchies have also been applied to summarization~\cite{ christensen2014hierarchical, zhang2017, tauchmann2018beyond}.

\section{Future Work}
\label{sec:future}

While \sectionref{shortcomings} focused on shortcomings in our modeling process and steps to fix them, this section focuses on broader guidelines for future work involving the \squash~format and its associated text generation task.

\paragraph{Evaluation of the \squash~format:} As discussed in \sectionref{introduction}, previous research shows support for the usefulness of hierarchies and QA in pedagogical applications. We did not directly evaluate this claim in the context of \squash, focusing instead on  evaluating the quality of QA pairs and their hierarchies. Moving forward, careful user studies are needed to evaluate the efficacy of the \squash~format in pedagogical applications, which might be heavily domain-dependent; for example, a QA hierarchy for a research paper is likely to be more useful to an end user than a QA hierarchy for an online blog. An important caveat is the imperfection of modern text generation systems, which might cause users to prefer the original human-written document over a generated \squash~output. One possible solution is a three-way comparison between the original document, a human-written \squash ed document, and a system-generated output. For fair comparison, care should be taken to prevent experimenter bias while crowdsourcing QA hierarchies (e.g., by maintaining similar text complexity in the two human-written formats).

\paragraph{Collection of a \squash~dataset:} Besides measuring the usefulness of the QA hierarchies, a large dedicated dataset can help to facilitate end-to-end modeling. While asking human annotators to write full \squash ed documents will be expensive, a more practical option is to ask them to pair \general~and \specific~questions in our dataset to form meaningful hierarchies and write extra questions whenever no such pair exists.

\paragraph{QA budget and deeper specificity hierarchies:}
\label{sec:qabudget}
In our work, we generate questions for every sentence and filter bad questions with fixed thresholds. An alternative formulation is an adaptive model dependent on a user-specified QA budget, akin to ``target length'' in summarization systems, which would allow end users to balance coverage and brevity themselves.
A related modification is increasing the depth of the hierarchies. While two-level QA trees are likely sufficient for documents structured into short and focused paragraphs, deeper hierarchies can be useful for long unstructured chunks of text. Users can control this property via a ``maximum children per QA node'' hyperparameter, which along with the QA budget will determine the final depth of the hierarchy.

\section{Conclusion}
\label{sec:conclusion}
We propose \squash, a novel text generation task which converts a document into a hierarchy of QA pairs. We present and evaluate a system which leverages existing reading comprehension datasets to attempt solving this task. We believe \squash~is a challenging text generation task and we hope the community finds it useful to benchmark systems built for document understanding, question generation and question answering. Additionally, we hope that our specificity-labeled reading comprehension dataset is useful in other applications such as 1) finer control over question generation systems used in education applications, curiosity-driven chatbots and healthcare~\cite{du2017learning}.
\section*{Acknowledgements}
\label{sec:acknowledge}

We thank the anonymous reviewers for their insightful comments. In addition, we thank Nader Akoury, Ari Kobren, Tu Vu and the other members of the UMass NLP group for helpful comments on earlier drafts of the paper and suggestions on the paper's presentation. This work was supported in part by research awards from the Allen Institute for Artificial Intelligence and Adobe Research. 

\bibliography{acl2019}

\begin{thebibliography}{56}
\expandafter\ifx\csname natexlab\endcsname\relax\def\natexlab#1{#1}\fi

\bibitem[{Bahdanau et~al.(2015)Bahdanau, Cho, and
  Bengio}]{Bahdanau2014NeuralMT}
Dzmitry Bahdanau, Kyunghyun Cho, and Yoshua Bengio. 2015.
\newblock Neural machine translation by jointly learning to align and
  translate.
\newblock In \emph{Proc.~International Conference on Learning Representations
  (ICLR)}.

\bibitem[{Baumel et~al.(2018)Baumel, Eyal, and Elhadad}]{baumel2018query}
Tal Baumel, Matan Eyal, and Michael Elhadad. 2018.
\newblock Query focused abstractive summarization: Incorporating query
  relevance, multi-document coverage, and summary length constraints into
  seq2seq models.
\newblock \emph{arXiv preprint arXiv:1801.07704}.

\bibitem[{Brandt et~al.(2011)Brandt, Joachims, Yue, and
  Bank}]{brandt2011dynamic}
Christina Brandt, Thorsten Joachims, Yisong Yue, and Jacob Bank. 2011.
\newblock Dynamic ranked retrieval.
\newblock In \emph{Proceedings of the fourth ACM international conference on
  Web search and data mining}.

\bibitem[{Buyukkokten et~al.(2001)Buyukkokten, Garcia-Molina, and
  Paepcke}]{buyukkokten2001seeing}
Orkut Buyukkokten, Hector Garcia-Molina, and Andreas Paepcke. 2001.
\newblock Seeing the whole in parts: text summarization for web browsing on
  handheld devices.
\newblock In \emph{Proceedings of the 10th international conference on World
  Wide Web}.

\bibitem[{Choi et~al.(2018)Choi, He, Iyyer, Yatskar, tau Yih, Choi, Liang, and
  Zettlemoyer}]{ChoiQuAC2018}
Eunsol Choi, He~He, Mohit Iyyer, Mark Yatskar, Wen tau Yih, Yejin Choi, Percy
  Liang, and Luke Zettlemoyer. 2018.
\newblock Quac: Question answering in context.
\newblock In \emph{Proc.~Conference on Empirical Methods in Natural Language
  Processing ({EMNLP})}.

\bibitem[{Christensen et~al.(2014)Christensen, Soderland, Bansal
  et~al.}]{christensen2014hierarchical}
Janara Christensen, Stephen Soderland, Gagan Bansal, et~al. 2014.
\newblock Hierarchical summarization: Scaling up multi-document summarization.
\newblock In \emph{Proc.~Association for Computational Linguistics (ACL)}.

\bibitem[{Dang(2005)}]{dang2005overview}
Hoa~Trang Dang. 2005.
\newblock Overview of duc 2005.
\newblock In \emph{Document Understanding Conferences}.

\bibitem[{Davenport et~al.(1998)Davenport, De~Long, and
  Beers}]{davenport1998successful}
Thomas~H Davenport, David~W De~Long, and Michael~C Beers. 1998.
\newblock Successful knowledge management projects.
\newblock \emph{Sloan management review}, 39(2):43--57.

\bibitem[{Devlin et~al.(2019)Devlin, Chang, Lee, and
  Toutanova}]{devlin2019bert}
Jacob Devlin, Ming-Wei Chang, Kenton Lee, and Kristina Toutanova. 2019.
\newblock Bert: Pre-training of deep bidirectional transformers for language
  understanding.
\newblock In \emph{Proc.~Conference of the North {A}merican Chapter of the
  Association for Computational Linguistics -- Human Language Technologies
  ({NAACL HLT})}.

\bibitem[{Du and Cardie(2017)}]{du2017identifying}
Xinya Du and Claire Cardie. 2017.
\newblock Identifying where to focus in reading comprehension for neural
  question generation.
\newblock In \emph{Proc.~Conference on Empirical Methods in Natural Language
  Processing ({EMNLP})}.

\bibitem[{Du and Cardie(2018)}]{Du2018HarvestingPQ}
Xinya Du and Claire Cardie. 2018.
\newblock Harvesting paragraph-level question-answer pairs from wikipedia.
\newblock In \emph{Proc.~Association for Computational Linguistics (ACL)}.

\bibitem[{Du et~al.(2017)Du, Shao, and Cardie}]{du2017learning}
Xinya Du, Junru Shao, and Claire Cardie. 2017.
\newblock Learning to ask: Neural question generation for reading
  comprehension.
\newblock In \emph{Proc.~Association for Computational Linguistics (ACL)}.

\bibitem[{Fan et~al.(2018)Fan, Lewis, and Dauphin}]{Fan2018HierarchicalNS}
Angela Fan, Mike Lewis, and Yann Dauphin. 2018.
\newblock Hierarchical neural story generation.
\newblock In \emph{Proc.~Association for Computational Linguistics (ACL)}.

\bibitem[{Gao et~al.(2018)Gao, Wang, Bing, King, and Lyu}]{gao2018difficulty}
Yifan Gao, Jianan Wang, Lidong Bing, Irwin King, and Michael~R Lyu. 2018.
\newblock Difficulty controllable question generation for reading
  comprehension.
\newblock \emph{arXiv preprint arXiv:1807.03586}.

\bibitem[{Gardner et~al.(2018)Gardner, Grus, Neumann, Tafjord, Dasigi, Liu,
  Peters, Schmitz, and Zettlemoyer}]{gardner2018allennlp}
Matt Gardner, Joel Grus, Mark Neumann, Oyvind Tafjord, Pradeep Dasigi, Nelson
  Liu, Matthew Peters, Michael Schmitz, and Luke Zettlemoyer. 2018.
\newblock Allennlp: A deep semantic natural language processing platform.
\newblock \emph{arXiv preprint arXiv:1803.07640}.

\bibitem[{Gruber(2008)}]{gruber2008collective}
Tom Gruber. 2008.
\newblock Collective knowledge systems: Where the social web meets the semantic
  web.
\newblock \emph{Web semantics: science, services and agents on the World Wide
  Web}, 6(1).

\bibitem[{Hakkarainen and Sintonen(2002)}]{hakkarainen2002interrogative}
Kai Hakkarainen and Matti Sintonen. 2002.
\newblock The interrogative model of inquiry and computer-supported
  collaborative learning.
\newblock \emph{Science \& Education}, 11(1):25--43.

\bibitem[{Hardtke et~al.(2009)Hardtke, Wertheim, and
  Cramer}]{hardtke2009demonstration}
David Hardtke, Mike Wertheim, and Mark Cramer. 2009.
\newblock Demonstration of improved search result relevancy using real-time
  implicit relevance feedback.
\newblock \emph{Understanding the User-Logging and Interpreting User
  Interactions in Information Search and Retrieval (UIIR-2009)}.

\bibitem[{Heilman and Smith(2010)}]{heilman2010good}
Michael Heilman and Noah~A Smith. 2010.
\newblock Good question! statistical ranking for question generation.
\newblock In \emph{Proc.~Conference of the North {A}merican Chapter of the
  Association for Computational Linguistics -- Human Language Technologies
  ({NAACL HLT})}.

\bibitem[{Henderson et~al.(2018)Henderson, Islam, Bachman, Pineau, Precup, and
  Meger}]{Henderson2018DeepRL}
Peter Henderson, Riashat Islam, Philip Bachman, Joelle Pineau, Doina Precup,
  and David Meger. 2018.
\newblock Deep reinforcement learning that matters.
\newblock In \emph{Proc.~Association for the Advancement of Artificial
  Intelligence (AAAI)}.

\bibitem[{Hintikka(1981)}]{hintikka1981logic}
Jaakko Hintikka. 1981.
\newblock The logic of information-seeking dialogues: A model.
\newblock \emph{Werner Becker and Wilhelm K. Essler Konzepte der Dialektik},
  pages 212--231.

\bibitem[{Hintikka(1988)}]{hintikka1988logic}
Jaakko Hintikka. 1988.
\newblock What is the logic of experimental inquiry?
\newblock \emph{Synthese}, 74(2):173--190.

\bibitem[{Hochreiter and Schmidhuber(1997)}]{hochreiter1997long}
Sepp Hochreiter and J{\"u}rgen Schmidhuber. 1997.
\newblock Long short-term memory.
\newblock \emph{Neural computation}.

\bibitem[{Holtzman et~al.(2018)Holtzman, Buys, Forbes, Bosselut, Golub, and
  Choi}]{Holtzman2018LearningTW}
Ari Holtzman, Jan Buys, Maxwell Forbes, Antoine Bosselut, David Golub, and
  Yejin Choi. 2018.
\newblock Learning to write with cooperative discriminators.
\newblock In \emph{Proc.~Association for Computational Linguistics (ACL)}.

\bibitem[{Holtzman et~al.(2019)Holtzman, Buys, Forbes, and
  Choi}]{holtzman2019curious}
Ari Holtzman, Jan Buys, Maxwell Forbes, and Yejin Choi. 2019.
\newblock The curious case of neural text degeneration.
\newblock \emph{arXiv preprint arXiv:1904.09751}.

\bibitem[{Honnibal and Montani(2017)}]{spacy2}
Matthew Honnibal and Ines Montani. 2017.
\newblock spacy 2: Natural language understanding with bloom embeddings,
  convolutional neural networks and incremental parsing.
\newblock \emph{To appear}.

\bibitem[{Jain et~al.(2018)Jain, Lazebnik, and Schwing}]{jain2018two}
Unnat Jain, Svetlana Lazebnik, and Alexander~G Schwing. 2018.
\newblock Two can play this game: visual dialog with discriminative question
  generation and answering.
\newblock In \emph{Proc.~IEEE Computer Society Conf. Computer Vision and
  Pattern Recognition (CVPR)}.

\bibitem[{Kim et~al.(2018)Kim, Lee, Shin, and Jung}]{kim2018improving}
Yanghoon Kim, Hwanhee Lee, Joongbo Shin, and Kyomin Jung. 2018.
\newblock Improving neural question generation using answer separation.
\newblock \emph{arXiv preprint arXiv:1809.02393}.

\bibitem[{Kim(2014)}]{kim2014convolutional}
Yoon Kim. 2014.
\newblock Convolutional neural networks for sentence classification.
\newblock In \emph{Proc.~Conference on Empirical Methods in Natural Language
  Processing ({EMNLP})}.

\bibitem[{Kingma and Ba(2014)}]{Kingma2014AdamAM}
Diederik~P. Kingma and Jimmy Ba. 2014.
\newblock Adam: A method for stochastic optimization.
\newblock In \emph{Proc.~International Conference on Learning Representations
  (ICLR)}.

\bibitem[{Landis and Koch(1977)}]{landis1977measurement}
J~Richard Landis and Gary~G Koch. 1977.
\newblock The measurement of observer agreement for categorical data.
\newblock \emph{biometrics}, pages 159--174.

\bibitem[{Lehnert(1978)}]{lehnert1978process}
Wendy~G Lehnert. 1978.
\newblock \emph{The process of question answering: A computer simulation of
  cognition}, volume 978.
\newblock Lawrence Erlbaum Hillsdale, NJ.

\bibitem[{Luong et~al.(2015)Luong, Pham, and Manning}]{luong2015effective}
Thang Luong, Hieu Pham, and Christopher~D Manning. 2015.
\newblock Effective approaches to attention-based neural machine translation.
\newblock In \emph{Proc.~Conference on Empirical Methods in Natural Language
  Processing ({EMNLP})}.

\bibitem[{Marchionini(2006)}]{marchionini2006exploratory}
Gary Marchionini. 2006.
\newblock Exploratory search: from finding to understanding.
\newblock \emph{Communications of the ACM}, 49(4):41--46.

\bibitem[{Paszke et~al.(2017)Paszke, Gross, Chintala, Chanan, Yang, DeVito,
  Lin, Desmaison, Antiga, and Lerer}]{paszke2017automatic}
Adam Paszke, Sam Gross, Soumith Chintala, Gregory Chanan, Edward Yang, Zachary
  DeVito, Zeming Lin, Alban Desmaison, Luca Antiga, and Adam Lerer. 2017.
\newblock Automatic differentiation in pytorch.
\newblock In \emph{NIPS 2017 Autodiff Workshop}.

\bibitem[{Peters et~al.(2018)Peters, Neumann, Iyyer, Gardner, Clark, Lee, and
  Zettlemoyer}]{peters2018deep}
Matthew Peters, Mark Neumann, Mohit Iyyer, Matt Gardner, Christopher Clark,
  Kenton Lee, and Luke Zettlemoyer. 2018.
\newblock Deep contextualized word representations.
\newblock In \emph{Proc.~Conference of the North {A}merican Chapter of the
  Association for Computational Linguistics -- Human Language Technologies
  ({NAACL HLT})}.

\bibitem[{Radford et~al.(2019)Radford, Wu, Child, Luan, Amodei, and
  Sutskever}]{radford2019language}
Alec Radford, Jeff Wu, Rewon Child, David Luan, Dario Amodei, and Ilya
  Sutskever. 2019.
\newblock Language models are unsupervised multitask learners.

\bibitem[{Rajpurkar et~al.(2018)Rajpurkar, Jia, and
  Liang}]{Rajpurkar2018KnowWY}
Pranav Rajpurkar, Robin Jia, and Percy Liang. 2018.
\newblock Know what you don't know: Unanswerable questions for squad.
\newblock In \emph{Proc.~Association for Computational Linguistics (ACL)}.

\bibitem[{Rajpurkar et~al.(2016)Rajpurkar, Zhang, Lopyrev, and
  Liang}]{rajpurkar2016squad}
Pranav Rajpurkar, Jian Zhang, Konstantin Lopyrev, and Percy Liang. 2016.
\newblock Squad: 100,000+ questions for machine comprehension of text.
\newblock In \emph{Empirical Methods in Natural Language Processing}, pages
  2383--2392.

\bibitem[{Reddy et~al.(2018)Reddy, Chen, and Manning}]{reddy2018coqa}
Siva Reddy, Danqi Chen, and Christopher~D. Manning. 2018.
\newblock Coqa: A conversational question answering challenge.
\newblock \emph{arXiv preprint arXiv:1808.07042}.

\bibitem[{Sachan and Xing(2018)}]{sachan2018self}
Mrinmaya Sachan and Eric Xing. 2018.
\newblock Self-training for jointly learning to ask and answer questions.
\newblock In \emph{Proc.~Conference of the North {A}merican Chapter of the
  Association for Computational Linguistics -- Human Language Technologies
  ({NAACL HLT})}.

\bibitem[{Schank(1972)}]{schank1972conceptual}
Roger~C Schank. 1972.
\newblock Conceptual dependency: A theory of natural language understanding.
\newblock \emph{Cognitive psychology}, 3(4):552--631.

\bibitem[{See et~al.(2017)See, Liu, and Manning}]{see2017get}
Abigail See, Peter~J Liu, and Christopher~D Manning. 2017.
\newblock Get to the point: Summarization with pointer-generator networks.
\newblock In \emph{Proc.~Association for Computational Linguistics (ACL)}.

\bibitem[{Sennrich et~al.(2016)Sennrich, Haddow, and
  Birch}]{sennrich2016neural}
Rico Sennrich, Barry Haddow, and Alexandra Birch. 2016.
\newblock Neural machine translation of rare words with subword units.
\newblock In \emph{Proc.~Association for Computational Linguistics (ACL)}.

\bibitem[{Serban et~al.(2016)Serban, Garc{\'\i}a-Dur{\'a}n, Gulcehre, Ahn,
  Chandar, Courville, and Bengio}]{serban2016generating}
Iulian~Vlad Serban, Alberto Garc{\'\i}a-Dur{\'a}n, Caglar Gulcehre, Sungjin
  Ahn, Sarath Chandar, Aaron Courville, and Yoshua Bengio. 2016.
\newblock Generating factoid questions with recurrent neural networks: The 30m
  factoid question-answer corpus.
\newblock In \emph{Proc.~Association for Computational Linguistics (ACL)}.

\bibitem[{Stede and Schlangen(2004)}]{stede2004information}
Manfred Stede and David Schlangen. 2004.
\newblock Information-seeking chat: Dialogues driven by topic-structure.
\newblock In \emph{Proceedings of Catalog (the 8th workshop on the semantics
  and pragmatics of dialogue; SemDial04)}.

\bibitem[{Tang et~al.(2017)Tang, Duan, Qin, Yan, and Zhou}]{tang2017question}
Duyu Tang, Nan Duan, Tao Qin, Zhao Yan, and Ming Zhou. 2017.
\newblock Question answering and question generation as dual tasks.
\newblock \emph{arXiv preprint arXiv:1706.02027}.

\bibitem[{Tauchmann et~al.(2018)Tauchmann, Arnold, Hanselowski, Meyer, and
  Mieskes}]{tauchmann2018beyond}
Christopher Tauchmann, Thomas Arnold, Andreas Hanselowski, Christian~M Meyer,
  and Margot Mieskes. 2018.
\newblock Beyond generic summarization: A multi-faceted hierarchical
  summarization corpus of large heterogeneous data.
\newblock In \emph{Proceedings of the Eleventh International Conference on
  Language Resources and Evaluation (LREC)}.

\bibitem[{Vaswani et~al.(2017)Vaswani, Shazeer, Parmar, Uszkoreit, Jones,
  Gomez, Kaiser, and Polosukhin}]{Vaswani2017AttentionIA}
Ashish Vaswani, Noam Shazeer, Niki Parmar, Jakob Uszkoreit, Llion Jones,
  Aidan~N. Gomez, Lukasz Kaiser, and Illia Polosukhin. 2017.
\newblock Attention is all you need.
\newblock In \emph{Proc.~Neural Information Processing Systems ({NIPS})}.

\bibitem[{Wagner(2004)}]{wagner2004wiki}
Christian Wagner. 2004.
\newblock Wiki: A technology for conversational knowledge management and group
  collaboration.
\newblock \emph{Communications of the association for information systems},
  13(1):19.

\bibitem[{Wagner and Bolloju(2005)}]{wagner2005supporting}
Christian Wagner and Narasimha Bolloju. 2005.
\newblock Supporting knowledge management in organizations with conversational
  technologies: Discussion forums, weblogs, and wikis.
\newblock \emph{Journal of Database Management}, 16(2).

\bibitem[{Wang et~al.(2017)Wang, Yuan, and Trischler}]{wang2017joint}
Tong Wang, Xingdi Yuan, and Adam Trischler. 2017.
\newblock A joint model for question answering and question generation.
\newblock \emph{arXiv preprint arXiv:1706.01450}.

\bibitem[{Wolf et~al.(2019)Wolf, Sanh, Chaumond, and Delangue}]{wolf2019}
Thomas Wolf, Victor Sanh, Julien Chaumond, and Clement Delangue. 2019.
\newblock Transfertransfo: {A} transfer learning approach for neural network
  based conversational agents.
\newblock \emph{CoRR}, abs/1901.08149.

\bibitem[{Yatskar(2019)}]{yatskar2018qualitative}
Mark Yatskar. 2019.
\newblock A qualitative comparison of coqa, squad 2.0 and quac.
\newblock \emph{Proc.~Human Language Technology/Conference of the North
  {A}merican Chapter of the Association for Computational Linguistics
  ({HLT/NAACL})}.

\bibitem[{Yee et~al.(2003)Yee, Swearingen, Li, and Hearst}]{yee2003faceted}
Ka-Ping Yee, Kirsten Swearingen, Kevin Li, and Marti Hearst. 2003.
\newblock Faceted metadata for image search and browsing.
\newblock In \emph{Proceedings of the SIGCHI conference on Human factors in
  computing systems}.

\bibitem[{Zhang et~al.(2017)Zhang, Verou, and Karger}]{zhang2017}
Amy~X. Zhang, Lea Verou, and David Karger. 2017.
\newblock Wikum: Bridging discussion forums and wikis using recursive
  summarization.
\newblock In \emph{Conference on Computer Supported Cooperative Work and Social
  Computing (CSCW)}.

\end{thebibliography}
\bibliographystyle{acl_natbib}

\newpage
\appendix

\setcounter{table}{0} \renewcommand{\thetable}{A\arabic{table}} 
\setcounter{figure}{0} \renewcommand{\thefigure}{A\arabic{figure}} 

\section*{Appendix}

\section{Question Classification Details}
\label{appendix:classification}
Confirming our intuition, \tableref{class_distro} shows us that QuAC has the highest percentage of \general~ questions. On the other hand CoQA and SQuAD, which allowed the question-asker to look at the passage, are dominated by \specific~ questions. These findings are consistent with a comparison across the three datasets in~\newcite{yatskar2018qualitative}. Interestingly, the average answer length for \specific~ questions in QuAC is 12 tokens, compared to 17 tokens for \general~ questions.\\
We provide the exact distribution of rule-labeled, hand-labeled and classifier-labeled questions in \tableref{scheme_distro}.\\

\section{Hyperparameters for Question Generation}
\label{appendix:hyperparameters}
Our question generation system consists of a two layer bidirectional LSTM encoder and a unidirectional LSTM decoder respectively. The LSTM hidden unit size in each direction and token embedding size is each set to 512. The class specificity embeddings size is 16. Embeddings are shared between the paragraph encoder and question decoder. All attention computations use a bilinear product~\citep{luong2015effective}. A dropout of 0.5 is used between LSTM layers. Models are trained using Adam~\cite{Kingma2014AdamAM} with a learning rate of $10^{-3}$, with a gradient clipping of 5.0 and minibatch size 32. Early stopping on validation perplexity is used to choose the best question generation model.

\section{What did not work?}
\label{appendix:notwork}
\paragraph{End-to-End Sequential Generation.} We experimented with an end-to-end neural model which generated a sequence of questions given a sequence of answer spans. As training data, we leveraged the sequence IDs and follow-up information in the QuAC dataset, \emph{without} specificity labels. We noticed that during decoding the model rarely attended over the history and often produced questions irrelevant to the context. A potential future direction would involve using the specificity labels for an end-to-end model.

\paragraph{Span Selection NER system.} As discussed in \sectionref{model_span} and~\newcite{du2017identifying}, we could frame answer span selection as a sequence labelling problem. We experimented with the NER system in AllenNLP (with ELMo embeddings) on the QuAC dataset, after the ground truth answer spans marked with BIO tags, after overlapping answers were merged together. We recorded low F1 scores of 33.3 and 15.6 on sentence-level and paragraph-level input respectively.
\paragraph{Paragraph-level question generation.} Our question generation model rarely generated \general~questions representative of the entire paragraph, even when we fed the entire paragraph as the answer span. We noticed that most \general~questions in our dataset were answered by one or two sentences in the paragraph.

\paragraph{Answering system trained on all datasets.} Recently,~\newcite{yatskar2018qualitative} reported small improvements on the QuAC validation set by pre-training the BiDAF++ model on SQuAD 2.0 or CoQA. We tried combining the training data in all three datasets but achieved a validation F1 score of just 29.3 (compared to 50.2 after using just QuAC training data).

\begin{table}[h!]
\begin{center}
\begin{tabular}{ lrrrr } 
 \toprule
Dataset & Size & Rule & Hand & CNN \\ 
\midrule
SQuAD & 86.8k & 30.5\% & 0.0\% & 69.5\% \\
QuAC & 65.2k & 59.3\% & 1.5\% & 39.2\% \\
CoQA & 105.6k & 57.1\% & 0.1\% & 42.8\% \\
All & 257.6k & 48.7\% & 0.4\% & 50.9\% \\
\bottomrule
\end{tabular}
\end{center}
\caption{Distribution of scheme adopted to classify questions in different datasets. ``CNN'' refers to the data-driven classifier. Roughly half the questions were classified using the rules described in \tableref{templates}.}
\label{tab:scheme_distro}
\end{table}

\begin{figure*}[t!]
\centering
\includegraphics[scale=0.97]{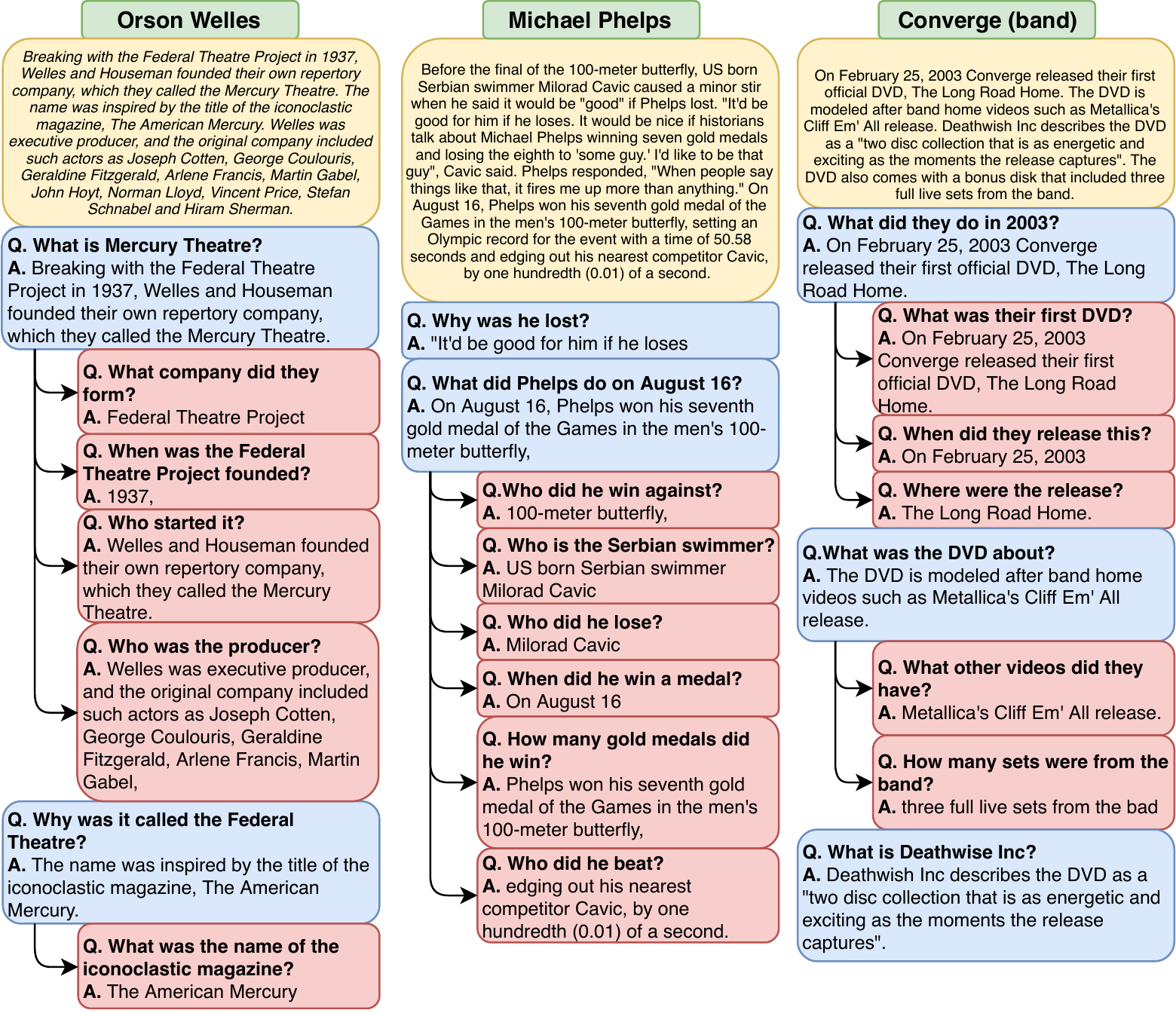}
\caption{Three \squash~outputs generated by our system, showcasing the strengths and weaknesses described in \sectionref{qual}.}
\label{fig:mass_examples}
\end{figure*}

\newpage
~
\newpage 
~
\newpage
\section{Technical Note on Modified System}
\label{appendix:modifications}

This technical note describes the modifications made to the originally published system to make a faster and more accurate system. We have incorporated language modelling pre-training in our modules using GPT-2 small~\cite{radford2019language} for question generation and BERT~\cite{devlin2019bert} for question answering. The official code and demo uses the modified version of the system.

\subsection{Dataset}
The primary modification in the dataset tackles the problem of coreferences in \general~questions, as described in \sectionref{shortcomings}. This is a common problem in QuAC and CoQA due to their contextual setup. We pass every question through the spaCy pipeline extension \texttt{neuralcoref}\footnote{\url{https://github.com/huggingface/neuralcoref/}} using the paragraph context to resolve co-references. We have also black-listed a few more question templates (such as ``\textit{What happened in <year>?}'') due to their unusually high prevalence in the dataset.

\subsection{Question Generation}
Our question generation system is now fine-tuned from a pretrained GPT-2 small model~\cite{radford2019language}. Our modified system is based on~\newcite{wolf2019} and uses their codebase\footnote{\url{https://github.com/huggingface/transfer-learning-conv-ai}} as a starting point.

\begin{figure}
\centering
\includegraphics[scale=0.9]{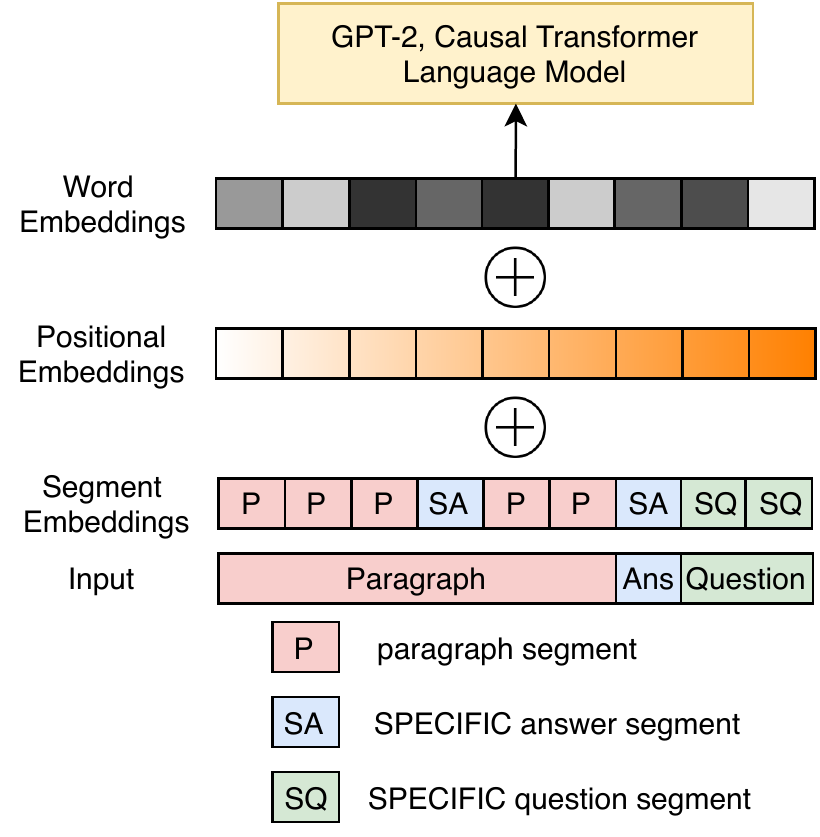}
\caption{An illustration of the model used for generating a \specific~question. A paragraph (context), answer and question and concatenated and the model is optimized to generate the question. Separate segment embeddings are used for paragraphs, \general~answers, \general~questions, \specific~answers and \specific~questions. Note that the answer segment embedding is also used within the paragraph segment to denote the location of the answer.}
\label{fig:squash_technical}
\end{figure}

We train our question generation model using the paragraph and answer as language modelling context. For \general~questions, our input sequence looks like \texttt{``<bos> ..paragraph text.. <answer-general> ..answer text.. <question-general> ..question text.. <eos>''} and equivalently for \specific~questions. In addition, we leverage GPT-2's segment embeddings to denote the specificity of the answer and question. Each token in the input is assigned one out of five segment embeddings (paragraph, \general~answer, \specific~answer, \general~question and \specific~question). Finally, answer segment embeddings were used in place of paragraph segment embeddings at the location of the answer in the paragraph to denote the position of the answer in the paragraph. For an illustration, refer to \figureref{squash_technical}.

The question generation model now uses top-$p$ nucleus sampling with $p = 0.9$~\cite{holtzman2019curious} instead of beam search and top-$k$ sampling. Due to improved question generation quality, we no longer need to over-generate questions.

\subsection{Question Answering}
We have switched to a BERT-based question answering module~\cite{devlin2019bert} which is trained on SQuAD 2.0~\cite{Rajpurkar2018KnowWY}. We have used an open source PyTorch implementation to train this model\footnote{\url{https://github.com/huggingface/pytorch-pretrained-BERT}}.

\subsection{Question Filtering}
We have simplified the question filtering process to incorporate a simple QA budget (described in \sectionref{qabudget}). Users are allowed to specify a custom ``\general~fraction'' and ``\specific~fraction'' which denotes the fraction of \general~and \specific~questions retained in the final output.
\end{document}